# Enhancing kidney transplantation through multi-agent kidney exchange programs: A comprehensive review and optimization models

**Shayan Sharifi**[a*]

[a]Wayne State University, United States

| C H R O N I C L E | A B S T R A C T |
|---|---|
|  | This paper presents a comprehensive review of the last two decades of research on Kidney Exchange Programs (KEPs), systematically categorizing and classifying key contributions to provide readers with a structured understanding of advancements in the field. The review highlights the evolution of KEP methodologies and lays the foundation for our contribution. We propose three mathematical models aimed at improving both the quantity and quality of kidney transplants. Model 1 maximizes the number of transplants by focusing on compatibility based on blood type and PRA, without additional constraints. Model 2 introduces a minimum Human Leukocyte Antigen (HLA) compatibility threshold to enhance transplant quality, though this leads to fewer matches. Model 3 extends the problem to a Multi-Agent Kidney Exchange Program (MKEP), pooling incompatible donor-recipient pairs across multiple agents, resulting in a higher number of successful transplants while ensuring fairness across agents. Sensitivity analyses demonstrate trade-offs between transplant quantity and quality, with Model 3 striking the optimal balance by leveraging multi-agent collaboration to improve both the number and quality of transplants. These findings underscore the potential benefits of more integrated kidney exchange systems. |
| |  |

## 1. Introduction

Kidney transplantation is one of the most effective treatments for kidney failure, involving the transfer of a kidney from a living or deceased donor to a patient in need (Li et al., 2016; Bay & Hebert, 1987). Historically, deceased donors were the primary source for transplants, but this approach only partially addressed the growing demand. Over time, the practice of using living donors has become increasingly prevalent, with innovations such as kidney division increasing the organ supply and saving more lives (Horvat et al., 2009). Additionally, research has shown that transplants from living donors are generally twice as successful as those from deceased donors, emphasizing the importance of expanding living donor programs. However, a significant barrier remains ensuring compatibility between donor and recipient (Dharia et al., 2022; Glorie et al., 2014).

In the United States, organ donation and transplantation have reached notable milestones. In 2023, over 46,000 organ transplants were performed, including more than 16,000 from deceased donors and nearly 7,000 from living donors. This marks a steady annual increase, with an average of 127 transplants occurring daily. Despite these achievements, over 103,000 individuals remain on the transplant waiting list, and tragically, 17 people die each day while waiting for an organ ("Organ Procurement and Transplantation Network," 2023). On a global scale, the disparity between the number of available organs and the demand remains acute. For example, in 2022, there were 102,090 kidney transplants and 37,436 liver transplants performed worldwide, with deceased donors accounting for approximately 41,792 transplants. Living donors, however, continue to play a pivotal role, particularly in countries with well-established programs ("United Network for Organ Sharing," 2023). Although the U.S. leads in both deceased and living organ donation, the shortage of organs remains a critical challenge worldwide.

A key issue in kidney transplantation is donor-recipient incompatibility, which occurs when a donor's kidney is unsuitable for the recipient due to blood or tissue mismatches. This poses a significant barrier to successful transplants. Paired kidney exchanges offer a solution to this problem, where two incompatible patient-donor pairs exchange kidneys with one another, thereby creating compatible matches. In cases where a perfect match cannot be found, the KEP allows for kidney exchanges between donor-recipient pairs with some degree of incompatibility, expanding transplant opportunities (Constantino et al., 2013). The goal of such programs is to maximize successful transplants through optimal compatibility matching. These exchanges can involve multiple pairs, with the simplest case involving two pairs where each donor is compatible with the other pair's recipient (Yuh et al., 2017). Rapaport (1986) was pioneer in developing the principles of paired kidney donation, envisioning two incompatible patient-donor pairs exchanging compatible kidneys. In this method, once two incompatible pairs are identified, they exchange kidneys, ensuring both patients receive compatible organs from the other pair's donor. This foundational concept of paired kidney exchange is at the core of our research.

Despite the success of KEPs, they face limitations in achieving optimal matches and managing logistical complexities. To address these issues, the concept of MKEPs has emerged as a promising alternative. While KEPs have been implemented at national and regional levels in many countries, recent initiatives in Europe seek to create international pools for MKEPs. This collaborative effort aims to increase the likelihood of finding compatible matches by combining the resources and donor pools of multiple nations (Ashlagi & Rot, 2014; Benedek et al., 2021; Mincu et al., 2021). By utilizing collaborative networks and advanced algorithms, MKEPs have the potential to transform the organ allocation process in the United States.

One of the key contributions of this paper is our focus on HLA compatibility to enhance transplant quality, particularly in cases where compatible matches are difficult to find. Initially, we implemented the general model in KEP. Then, by introducing minimum HLA compatibility requirements, we demonstrate that the number of transplants decreases. Consequently, we propose a final model that incorporates MKEP, taking HLA into account by considering fairness, ensuring all agents can receive at least the number of transplants they would obtain if acting independently. Our results show that with a larger pool of incompatible pairs, it is possible to simultaneously increase both the number and the quality of transplants. Further details are provided in the model and numerical example sections.

In this paper, we explore the potential benefits and feasibility of implementing MKEPs through a detailed numerical example. Our goal is to address critical gaps in transplant accessibility and efficiency, ultimately improving outcomes for patients in need of kidney transplants. Expanding the U.S. kidney exchange program through MKEPs could increase the number of transplants by 30 to 63 percent. However, current research highlights significant inefficiencies in the existing system, as most transplants are coordinated by individual hospitals rather than national platforms. This fragmentation leads to suboptimal outcomes, as hospitals often fail to fully consider the broader benefits of participating in exchanges, and existing platforms lack sufficient incentives for hospitals to submit patients and donors. Solving this problem requires a combination of new mechanisms, reforms, and reimbursement strategies (Agarwal et al., 2019).

To further support this argument, an evaluation of the National Kidney Registry (NKR), the largest kidney exchange network in the U.S., was conducted using data from the Scientific Registry of Transplant Recipients (SRTR). The SRTR provides comprehensive data on kidney donors, transplant candidates, and recipients. The analysis revealed that patients at hospitals affiliated with the NKR are 2.5 to 3 times more likely to receive a transplant from a living donor compared to those at non-affiliated hospitals, demonstrating the effectiveness of collaborative networks (Ghanbariamin & Chung, 2020).

This research addresses two key issues in kidney transplantation: the shortage of available organs and the quality of transplants. A major concern is the mortality rate among patients who die while waiting for transplants due to long waiting times. Another critical factor is the level of HLA compatibility between the patient and donor, which significantly influences the quality of the transplant. A higher HLA compatibility rate increases the likelihood of organ acceptance by the recipient, reducing the risk of rejection (Clark & Unsworth, 2010). Moreover, patients who receive high-quality transplants experience better long-term health outcomes. Thus, improving kidney transplantation processes not only addresses the organ shortage but also enhances the quality of life for transplant recipients.

Additionally, this paper presents a comprehensive review of the last two decades of research on KEPs, providing readers with a detailed and informed perspective on the evolution of kidney exchange practices. To the best of our knowledge, this extensive review offers a valuable foundation for understanding the future potential of KEPs and MKEPs in improving transplant outcomes globally.

## 2. Literature review

The body of research surrounding KEPs has expanded significantly, addressing key challenges such as maximizing the number of transplants, incorporating failure probabilities, optimizing logistics, and balancing costs. This section reviews contributions to the field, focusing on common themes that have shaped the development of KEP methodologies and solutions.

*2.1 Maximizing Transplant Opportunities*

One of the primary goals in KEP research is maximizing the number of successful transplants, particularly within pools of incompatible patient-donor pairs. Alvelos et al. (2019) proposed an integer programming model that accounts for the probability of failure during matching. They later relaxed this model into a linear programming (LP) framework to maximize the number of transplants possible in incompatible sets (Alvelos et al., 2015). In a follow-up study, expanded their approach by incorporating various types of cycles and chains, employing branch-and-price methods to further optimize transplant opportunities in failure-prone scenarios (Alvelos et al., 2019).

Similarly, Dickerson et al. (2016) introduced scalable KEP formulations, combining two innovative approaches to handle large-scale kidney exchanges. Their model addressed issues of mismatches, including age and weight differences, and resolved compatibility challenges through simulations using real-world data. Klimentova et al. (2014) also proposed a cycle decomposition model with dual objectives of maximizing transplant numbers while minimizing costs, offering an efficient solution for large-scale exchanges.

Yuh et al. (2017) strengthened these efforts by employing the Reformulation Linearization Technique (RLT) to develop a new integer programming model. This model systematically improved upon earlier methods by enhancing lower bounds and optimizing matching quality. Abraham et al. (2007) contributed scalable algorithms for national KEP markets, focusing on maximizing social welfare while addressing incompatibility on a large scale.

Li et al. (2014) introduced two integer programming (IP) formulations aimed at optimizing kidney exchange organizations. They proposed a novel approach for maximizing kidney allocation, considering the specific needs of recipient groups on the waiting list. Their work also introduced random characteristics into the management of KEP programs, enhancing the flexibility and applicability of the model.

*2.2 Addressing Uncertainty and Failure*

Another key challenge in KEPs involves handling uncertainty and potential failures in the matching process. Ahmadvand and Pishvaee (2018) pioneered a model based on Data Envelopment Analysis (DEA) that evaluates patient-donor efficiency under uncertain conditions. This approach incorporated both medical and non-medical factors, introducing fuzzy programming to make allocation decisions more flexible in dynamic healthcare environments. Ahmadvand and Pishvaee (2018) developed a two-phase stochastic programming (SP) model to address node and arc failures in kidney exchanges. Their model improves robustness by mitigating system breakdowns prior to implementation (Lee et al., 2018). Zheng et al. (2015) similarly introduced a Stochastic Minimum Cost Flow (SMCF) model to handle uncertain arc failures. Their method ensured that alternative paths were available for rerouting, increasing the reliability of the exchange process.

In their efforts to optimize kidney exchanges without limiting chain lengths, Anderson developed an algorithm that closely replicates real-world KEP scenarios. Their solution, designed for practical applications, maximized the number of transplants while optimizing logistical constraint (Anderson et al., 2015).

*2.3 Enhancing Efficiency and Reducing Costs*

Several researchers have focused on improving the efficiency of KEPs by reducing logistical and operational costs. Caruso and Daniele (2018) presented a network-based model designed to minimize total costs associated with transplantation, including hospital, surgery, and transportation expenses. By using non-linear formulations, their model ensured optimal resource distribution within national healthcare systems (Caruso & Daniele, 2018). Kutlu-Gundogdu et al. (2018) addressed the kidney transplant problem in Turkey with an Integer Programming (IP) approach, considering the demographic impact of age and gender on transplant outcomes. They demonstrated that demographic factors can play a significant role in optimizing the allocation process and improving efficiency.

Zahiri et al. (2014) introduced a robust probabilistic model for organ allocation, focusing on minimizing costs and waiting time while enhancing network performance. Their later work expanded this approach to include a dynamic location-allocation problem, ensuring that organs were allocated efficiently within transplant centers while keeping costs and waiting times low. Savaser et al. (2019) also emphasized operational efficiency by reducing transportation time between donor and recipient cities. By improving the logistics of organ transfer, their model enhanced surgical performance and increased the probability of successful transplants.

Caurso and Daniele (2018) presented a network-based mathematical model designed to minimize the total costs associated with organ transplantation, including expenses related to hospitals, surgeries, transportation, medical teams, and disposal. Their model aimed to optimize healthcare services by developing a diverse formula that accounts for unpredictable changes, providing solutions for cost reduction and operational efficiency. The study demonstrated that this model effectively minimized total costs while maintaining balance and efficiency across the healthcare system.

Zahiri et al. (2014) introduced a dynamic location-allocation problem for organ allocation under uncertainty for transplant centers (TC) units. Their model used a mixed-integer mathematical programming approach with dual objectives: optimizing the prioritization of organs and minimizing both costs and total waiting times for transplant surgeries. The model ensured that organs were allocated efficiently and in a timely manner, addressing both operational and logistical challenges in organ transplantation.

*2.4 Methodological Innovations and Multi-Criteria Appro*

Several contributions have advanced the methodological foundations of KEP research, with a focus on incorporating multi-criteria decision-making and innovative algorithms. Constantino et al. (2013) introduced two formulas—Lagrangian relaxation and strong compatibility limits—that have been widely adopted for handling compatibility constraints in complex kidney exchanges. Their approach relaxed traditional compatibility constraints, allowing for more flexible matching (Constantino et al., 2013).

Pansart et al. (2018) developed a column generation approach to overcome the NP-hard nature of pricing problems in KEPs. Their methods ensured high-quality solutions within short timeframes, enhancing the overall logistics of transplant allocation.

Glorie et al. (2012) made significant contributions by demonstrating how large, multi-criteria kidney exchanges can be optimized using scalable algorithms. Their multi-stage hierarchical approach effectively smoothed large cycles and chains, improving the efficiency of resource allocation and fairness in the matching process (Glorie et al., 2012, 2014).

Dickerson expanded on these innovations by introducing new formulations with limited cycle lengths and linear programming relaxations. Their work, based on real-world data from the U.S. and U.K., outperformed existing models in managing long kidney exchange chains (Dickerson et al., 2016).

Li et al. (2019) conducted two significant studies on matching compatible pairs in exchanges. Their work introduced the Living Kidney Donor Profile Index (LKDPI) to evaluate donor profiles, and they developed a simulation model that allowed for the joint evaluation of compatibility and quality. This innovative approach improved the allocation of compatible and incompatible pairs in real-world transplant centers (Li et al., 2019). Further details and classifications of these studies can be found in Table 1.

**Table 1**
Summary of Literature on KEPs

| Author | Year | Single or Multi Objective Function | Objective Function | Methodology | Solution Method | Deterministic or Uncertainty | Model Approach | Period |
|---|---|---|---|---|---|---|---|---|
| Dickerson et al. | 2016 | Single | Maximizing the number of potential transplants | Integer Programming | Heuristic | Stochastic | Optimization | Single |
| Glorie et al. | 2014 | Single | Maximizing score in each transplant | Integer Programming | Exact | Deterministic | Optimization and Simulation | Single |
| Li et al. | 2019 | Multi | Maximizing number of transplantations and chance of acceptance | Linear Programming | Heuristic | Robust | Simulation | Single |
| Zheng et al. | 2015 | Multi | Min transportations costs and Max social welfare | Integer Programming | Exact | Stochastic | Optimization | Single |
| Glorie et al. | 2018 | Multi | Max number of transplants and maximizing the minimum obtained value in each source | Integer Programming | Exact | Robust | Optimization and Simulation | Single |
| Dickerson et al. | 2017 | Single | Maximizing the number of potential transplants | Mixed Integer Programming | Exact | Deterministic | Optimization | Single |
| Dickerson et al. | 2019 | Multi | Maximizing number and quality of transplants | TSP | Heuristic | Robust | Optimization | Single |
| Pansart et al. | 2019 | Single | Minimizing time in cycle | Integer Programming | Exact/Heuristic | Deterministic | Optimization | Single |
| Glorie et al. | 2013 | Multi | Maximizing number and fairness of transplants | Mixed Integer Programming | Exact | Deterministic | Simulation | Single |
| Klimentova et al. | 2014 | Multi | Minimizing the cycle length and maximizing number of transplants | Mixed Integer Programming | Heuristic | Deterministic | Optimization | Single |
| Abraham et al. | 2007 | Single | Maximizing social welfare | Integer Programming | Heuristic | Stochastic | Optimization | Single |
| Rees et al. | 2014 | Single | Maximizing number of transplants | Integer Programming | Exact | Stochastic | Optimization and Simulation | Single |
| Alvelos et al. | 2016 | Single | Maximizing number of transplants | Integer Programming | Exact | Stochastic | Optimization | Single |
| Pishvaee et al. | 2017 | Single | Minimizing Deviation | DEA | Exact | Fuzzy | Simulation | Single |

**Table 1**
Summary of Literature on KEPs (Continued)

| Author | Year | Single or Multi Objective Function | Objective Function | Methodology | Solution Method | Deterministic or Uncertainty | Model Approach | Period |
|---|---|---|---|---|---|---|---|---|
| Alvelos et al. | 2016 | Single | Maximizing expected number of transplants | Integer Programming | Heuristic | Stochastic | Optimization | Single |
| Constantino et al. | 2013 | Single | Maximizing exchanged weight | Integer Programming | Heuristic | Deterministic | Optimization and Simulation | Single |
| Carsu et al. | 2017 | Single | Minimizing transplant cost | Linear Programming | Heuristic | Deterministic | Optimization | Single |
| Anderson et al. | 2015 | Single | Maximizing number of transplants | Integer Programming | Heuristic | Deterministic | Optimization and Simulation | Single |
| lee et al. | 2018 | Multi | Maximizing exchanged weight in each cycle and minimizing rejection probability | Integer Programming | Heuristic | Stochastic | Optimization and Simulation | Single |
| Aktin et al. | 2018 | Single | Maximizing number of transplants | Integer Programming | Exact | Deterministic | Optimization | Single |
| Chung et al. | 2017 | Multi | Maximizing exchanged weight | Mixed Integer Programming | Heuristic | Deterministic | Optimization | Single |
| Yetis et al. | 2018 | Single | Maximizing compatibility in ischemic time | Mixed integer nonlinear programming | Heuristic | Deterministic | Optimization and Simulation | Multi |
| Zahiri et al. | 2014 | Single | Minimizing time and cost of transplant | Mixed integer nonlinear programming | Metaheuristic | Robust | Optimization | Multi |
| Zahiri et al. | 2014 | Multi | Considering priority for assigning kidney and Minimizing costs of transplants | Mixed integer programming | Metaheuristic | Fuzzy | Optimization | Multi |
| Kargar et al. | 2020 | Multi | Minimizing transports costs and minimizing rejection rate after transplant | Mixed integer programming | Exact | Fuzzy | Optimization | Multi |
| Belien et al. | 2011 | Single | Minimizing transportation time | Mixed integer programming | Exact | Deterministic | Optimization | Multi |

In summary, the literature surrounding KEPs has evolved from basic integer programming models to more advanced approaches incorporating stochastic programming, robust optimization, and multi-criteria decision-making. Key contributions have focused on maximizing transplant opportunities, improving efficiency, and reducing the risks associated with uncertainty and failures.

One of the key contributions of this paper is our focus on HLA compatibility to enhance transplant quality, particularly in cases where finding compatible matches is challenging. Initially, we implement a general model within the KEP framework. By introducing minimum HLA compatibility requirements, we demonstrate that the number of transplants decreases due to the stricter matching criteria. To address this, we propose a final model that incorporates MKEP, accounting for HLA compatibility while ensuring fairness. This ensures that all agents receive at least the number of transplants they would achieve if managing their own pool independently. Our results show that by increasing the pool size of incompatible pairs, both the number and quality of transplants can be improved simultaneously. Further details are provided in the model and numerical example sections.

## 3. Model

In this section, we present the mathematical formulations for addressing the kidney transplant assignment problem, structured into three distinct models. Each model builds on the previous one to improve transplant outcomes by introducing new objectives and constraints.

- **Model 1** represents the general version of the KEP, aimed at maximizing the number of transplantations while considering blood type and PRA compatibility between patients and donors.

- **Model 2** builds on the previous formulation by introducing a minimum HLA compatibility threshold to ensure that all assigned transplants meet a high standard of quality. This model evaluates the impact of enforcing stricter HLA requirements on the transplantation outcomes.

- **Model 3** extends the problem to MKEP, where multiple agents (e.g., hospitals or regions) collaborate. This model demonstrates that, by combining the pools of incompatible pairs across agents, not only can the number of successful transplantations increase, but the quality of the transplants, as measured by HLA compatibility, can also be improved. Additionally, this model guarantees that each agent receives at least as many transplants as it would have achieved independently in Model 1.

The primary objective across all models is to maximize the total compatibility score between patients and donors, accounting for key factors like HLA compatibility, blood type, and PRA type. Below, we define the common variables and constraints for all models, followed by the additional constraints specific to each model.

**Set:**

$I$: index for each pair which includes an incompatible patient and donor

$A$: index for each agent which we assume can be $\{1, 2, 3, 4\}$

$N_A$: the set of patient-donor pairs for agent A

**Parameters:**

$HLA_S^{ij}$: HLA score between patient i and donor j
$HLA_T^{ij}$: HLA score between patient i and donor j, as well as patient j and donor i.

$$HLA_T^{ij} = HLA_S^{ij} + HLA_S^{ji} \tag{1}$$

$L_{HLA}$: Minimum requirement for HLA score for a high-quality transplant

$c_{ij}$: 1, if patient i and donor j and patient j and donor i are compatible in terms of body tissue and blood type

**Decision Variables:**

$x_{ij}$: 1, if patient i receives kidney from donor j, and patient j receives kidney from donor i
$p_{ij}$: 1, if patient i and donor j satisfy the minimum HLA requirement, $L_{HLA}$

Model 1:

This model aims to maximize the number of transplants by considering blood type and PRA compatibility.

$$\max \sum_{i \in N_A} \sum_{j \in N_A, j > i} x_{ij} \tag{2}$$

$$x_{ij} \leq c_{ij} \qquad \forall\, i \in N_A, j \in N_A, i \neq j \tag{3}$$

$$x_{ij} = x_{ji} \qquad \forall\, i,j \in I \tag{4}$$

$$\sum_{j=1, j \neq i}^{|N_A|} x_{\min(i,j), \max(i,j)} \leq 1 \qquad \forall\, i \in N_A \tag{5}$$

$$x_{ij} \in \{0,1\} \tag{6}$$

The objective function (Eq. 1) maximizes the total number of matched pairs across all incompatible patient-donor pairs in the set $N_A$, ensuring that each pair is counted once. The compatibility between patient i and donor j is governed by the parameter $c_{ij}$, which is calculated based on blood type and PRA compatibility (Eq. 3). A match can only occur if both conditions are satisfied, meaning that $c_{ij} = 1$, otherwise $x_{ij}$ must be 0. The symmetry constraint (Eq. 4) ensures that if a match between patient i and donor j is made, the reciprocal match is also valid, enforcing consistency in the matching process. Each patient can only receive one kidney, which is ensured by limiting the sum of assigned pairs for each patient to at most one (Eq. 5). Finally, the decision variable $x_{ij}$ is binary, ensuring that a match is either made or not made (Eq. 6). Together, this formulation creates a structure where the model prioritizes maximizing the number of transplants while ensuring that all matches are biologically feasible, and that no patient receives more than one kidney.

The parameter $c_{ij}$ represents the compatibility between patient iii and donor j based on two key biological factors: **blood type compatibility** and **PRA compatibility**. Here's how $c_{ij}$ is calculated:

- **Blood Type Compatibility**: For a transplant to be successful, the donor and recipient must have compatible blood types. Compatibility is determined based on the following rules:
    - Type O can donate to any blood type (universal donor).
    - Type A can donate to A and AB.
    - Type B can donate to B and AB.
    - Type AB can donate only to AB.
- **PRA (Panel Reactive Antibody) Compatibility**: PRA measures the level of antibodies in the patient's blood that could react against the donor's tissue. A high PRA indicates that the patient is more likely to reject the kidney. Compatibility is generally easier if the patient has a lower PRA score or if the donor has a low antigen profile that is less likely to provoke an immune response.

Therefore, $c_{ij}$ is calculated as:

- $c_{ij} = 1$ if both blood type and PRA compatibility are satisfied between patient i and donor j.
- $c_{ij} = 0$ if either the blood type or PRA compatibility condition is violated.

The matrix $c_{ij}$ is precomputed before solving the model and used to determine which pairs can be considered for transplantation.

An important consideration is the level of compatibility between the patient and donor, particularly in terms of HLA matching, which plays a critical role in determining the success of the transplant. Higher HLA compatibility between the patient and donor reduces the likelihood of organ rejection, as the recipient's body is more likely to accept the transplant. Moreover, patients who undergo transplants with higher compatibility rates generally experience improved health outcomes and enhanced post-transplant quality of life. For this reason, in **Model 2**, we introduce the concept of HLA matching into the objective function and establish a minimum HLA requirement to ensure that all kidney transplants meet a certain standard of quality. By adding this criterion, we aim to not only maximize the number of transplants but also improve the overall quality of life for transplant recipients by reducing the risk of rejection and increasing the long-term success rate of the transplant.

Model 2:

Model 2 builds on Model 1 by introducing a minimum HLA score requirement to improve transplant quality. The goal is to maximize the number of transplants while ensuring each match meets the HLA compatibility threshold.

$$\max \sum_{i \in N_A} \sum_{j \in N_A, j > i} HLA_T^{ij} x_{ij} \qquad (7)$$

$$x_{ij} \leq c_{ij} \qquad \forall\, i \in N_A, j \in N_A, i \neq j \qquad (8)$$

$$HLA_S^{ij} \geq L_{HLA} p_{ij} \qquad \forall\, i \in N_A, j \in N_A, i \neq j \qquad (9)$$

$$x_{ij} \leq p_{ij} \qquad \forall\, i \in N_A, j \in N_A, i \neq j \qquad (10)$$

$$x_{ij} = x_{ji} \qquad \forall\, i, j \in I \qquad (11)$$

$$\sum_{j=1, j \neq i}^{|N_A|} x_{\min(i,j), \max(i,j)} \leq 1 \qquad \forall\, i \in N_A \qquad (12)$$

$$x_{ij}, p_{ij} \in \{0,1\} \qquad (13)$$

In **Model 2**, two new elements are introduced to ensure a minimum level of HLA compatibility for each transplant. The first new constraint (Eq. 9) ensures that the HLA compatibility score between patient i and donor j, represented by $HLA_S^{ij}$, must meet or exceed a predefined threshold $L_{HLA}$ for the match to be considered viable. To enforce this, the binary variable $p_{ij}$ is introduced, which equals 1 if the HLA compatibility requirement is satisfied, i.e., if $HLA_S^{ij} \geq L_{HLA}$. The second new constraint (Eq. 10) ensures that a match $x_{ij}$ can only occur if the minimum HLA requirement is met, meaning $x_{ij} \leq p_{ij}$. In other words, the match can only proceed if $p_{ij} = 1$, guaranteeing that the match satisfies the required level of HLA compatibility. These additions ensure that the model not only maximizes the number of transplants but also improves the quality of each transplant by focusing on high HLA compatibility.

Model 3:

The new version of **MKEP** model (Model 3) aims to address two primary challenges in kidney transplantation: the shortage of transplants and the quality of transplants. By combining kidney pools across multiple agents (e.g., hospitals or regions), the MKEP model significantly increases the number of potential transplants, as more incompatible donor-recipient pairs can be matched. Additionally, by incorporating HLA compatibility into the model, we ensure that not only is the quantity of transplants maximized, but also the quality, thereby improving post-transplantation outcomes and reducing the risk of rejection. This approach is essential to enhance the long-term quality of life for transplant recipients by ensuring better matching.

In **Model 3**, we extend the framework to a multi-agent setting, where several agents (such as hospitals or countries) contribute their incompatible donor-recipient pairs to a shared pool. The model guarantees that each agent will receive at least the number of transplants they would have obtained individually, while maximizing the overall number and quality of transplants. We need to modify the sets, parameters, and variables of the previous model and provide the mathematical model based on the new changes.

**Sets and Indices:**

- A: Set of agents (e.g., hospitals, regions).
- $N_A$: Set of incompatible pairs for agent a, where $a \in A$.
- i, j: Indices of incompatible donor-recipient pairs within agent a.
- n: Total number of pairs across all agents.

**Decision Variables:**

- $x_{ij}^{st}$: A binary decision variable that equals 1 if patient i from agent s is matched with donor j from agent t (either within or between agents), and 0 otherwise.
- $p_{ij}^{st}$: A binary variable that equals 1 if the match between patient i from agent s and donor j from agent t meets the HLA compatibility threshold.

$$\max \sum_{s \in A} \sum_{t \in A} \sum_{i \in N_A} \sum_{j \in N_{A'}} HLA_{T_{ij}}^{st} x_{ij}^{st} \tag{14}$$

$$x_{ij}^{st} \leq c_{ij}^{st} \qquad \forall\, i,j \in N_A, s,t \in A, s \neq t \tag{15}$$

$$x_{ij}^{st} \leq c_{ij}^{st} \qquad \forall\, i,j \in N_A, s,t \in A, s = t, i \neq j \tag{16}$$

$$HLA_{S_{ij}}^{st} \geq L_{HLA} p_{ij}^{st} \qquad \forall\, i,j \in N_A, s,t \in A, s \neq t \tag{17}$$

$$HLA_{S_{ij}}^{st} \geq L_{HLA} p_{ij}^{st} \qquad \forall\, i,j \in N_A, s,t \in A, s = t, i \neq j \tag{18}$$

$$x_{ij}^{st} \leq p_{ij}^{st} \qquad \forall\, i,j \in N_A, s,t \in A, s \neq t \tag{19}$$

$$x_{ij}^{st} \leq p_{ij}^{st} \qquad \forall\, i,j \in N_A, s,t \in A, s = t, i \neq j \tag{20}$$

$$x_{ij}^{st} = x_{ji}^{ts} \qquad \forall\, i,j \in N_A, s,t \in A \tag{21}$$

$$\sum_{s \in A} \sum_{j=1, j \neq i}^{|N_A|} x_{\min(i,j)}^{st} \leq 1 \qquad \forall\, i \in N_A, s \in A \tag{22}$$

$$\sum_{i \in N_A} \sum_{j \in N_A, j > i} x_{ij}^{ss} + \sum_{t \in A, s \neq t} \sum_{i \in N_A} \sum_{j \in N_{A'}} x_{ij}^{st} \geq M_A \qquad \forall\, s \in A \tag{23}$$

$$x_{ij}^{st}, p_{ij}^{st} \in \{0,1\} \tag{24}$$

The objective function (Eq. 14) maximizes the total HLA compatibility score for all matched pairs between agents. Equations (15) and (16) ensure that a match between patient i and donor j, either between agents (cross-agent) or within the same agent (intra-agent), is only allowed if they are compatible based on blood type and PRA, represented by $c_{ij}^{st}$. Equations (17) and (18) introduce a minimum HLA compatibility threshold, where $HLA_{S_{ij}}^{st}$ must be greater than or equal to the predefined threshold $L_{HLA}$. The binary variable $p_{ij}^{st}$ is used to enforce this condition, where $p_{ij}^{st} = 1$ if the HLA compatibility score meets the requirement. Furthermore, constraints (19) and (20) ensure that a match $x_{ij}^{st}$ between patient i from agent s and donor j from agent t can only occur if the HLA compatibility requirement is satisfied, meaning $x_{ij}^{st} \leq p_{ij}^{st}$. To maintain the consistency of matching, Eq. (21) enforces symmetry, ensuring that if patient i from agent s is matched with donor j from agent t, the reciprocal match $x_{ij}^{st}$ is also valid. Each patient can be matched with only one donor, either within the same agent or across agents, as ensured by Eq. (22). Finally, Eq. (23) guarantees that each agent receives at least as many transplants as it would have achieved independently, maintaining fairness in the collaborative multi-agent system. The binary decision variables $x_{ij}^{st}$ and $p_{ij}^{st}$ (Eq. 24) ensure that the model respects both the matching and HLA compatibility conditions. Through this multi-agent framework, the model not only increases the number of transplants but also improves their quality by focusing on HLA compatibility between patients and donors.

- The model ensures that both intra-agent and cross-agent matches maximize HLA compatibility, while respecting compatibility constraints based on blood type and PRA, so it guarantees the transplants quality.

- It guarantees fairness, as each agent is assured a minimum number of transplants, equivalent to what they would achieve independently.

- The binary variables $x_{ij}^{st}$ and $p_{ij}^{st}$ control the feasibility of matches and ensure that HLA compatibility is met for all transplants.

- By considering all the above points, this model provides more transplants for all agents with a better quality as compared to the scenario that each agent run its own pool, which you can see the detail comparisons and analysis in the next section.

## 4. Discussion

To evaluate the effectiveness of our kidney transplantation models, we conducted a numerical example using simulated data for incompatible donor-recipient pairs. This example simulates a MKEP, where multiple agents, such as hospitals or states, contribute their own pools of incompatible donor-recipient pairs to a shared exchange pool. We explored three distinct cases to demonstrate the impact of both the number of transplants and the quality of matches when considering HLA compatibility thresholds, while also ensuring fairness for all agents. Initially, we intended to work on a real case study in the US. However, due to the unavailability of data from at least four states to simulate the problem based on real-world conditions, we generated data using Python tools.

### 4.1 Numerical example

In this example, four agents contributed $n$ incompatible pairs to the kidney exchange pool. Each agent's pool of incompatible pairs was characterized by blood type, PRA compatibility, and HLA compatibility scores between donor-recipient pairs. The primary goal of this numerical example was to examine how incorporating HLA compatibility thresholds affects both the number and quality of transplants, and how introducing a multi-agent system increases the overall efficiency of kidney exchanges. The simulation followed these key steps:

1. **Random Data Generation**: Compatibility data for blood type, PRA, and HLA scores were randomly generated for each pair within the $n$ pairs contributed by each agent. Blood type and PRA compatibility were represented by binary values, where 1 indicated compatibility and 0 indicated incompatibility. HLA scores were randomly drawn from a predefined set of values, reflecting various levels of compatibility between donors and recipients.

2. **HLA Compatibility Threshold**: For scenarios where HLA compatibility was considered, a minimum threshold value $L_{HLA}$ was set to ensure that only high-quality transplants were allowed. Pairs with HLA compatibility scores below this threshold were deemed incompatible for transplantation.

**Case 1:** Maximizing the Number of Transplants without HLA Compatibility

In the first case, we aimed to maximize the number of transplants within each agent's pool without considering HLA compatibility. The matching was based solely on blood type and PRA compatibility. The following constraints were applied:

- Matches were only allowed if pairs were compatible with respect to blood type and PRA.

- Each patient could receive only one transplant.

- The objective was to maximize the number of successful transplants for each individual agent.

This case served as the baseline, prioritizing the quantity of transplants without regard for match quality in terms of HLA compatibility.

**Case 2:** Adding Minimum HLA Compatibility Requirement

In the second case, we introduced a minimum HLA compatibility threshold $L_{HLA}$ for each agent's kidney exchange pool. The model incorporated this threshold into the matching process, ensuring that only high-quality transplants, meeting the minimum HLA score, were allowed. Key changes in this case included:

- A constraint that required the HLA compatibility score between patient iii and donor j to exceed the threshold $L_{HLA}$.

- The blood type and PRA compatibility constraints from Case 1 were retained.

- The objective was to maximize both the number and quality of transplants, ensuring that all matches met the minimum HLA compatibility requirement.

**Case 3:** Multi-Agent Kidney Exchange with HLA Compatibility

In the third case, the incompatible pairs from all four agents were combined into a shared pool, creating a multi-agent kidney exchange model. This allowed for cross-agent matching, which leveraged the larger pool of donor-recipient pairs to increase the number of transplants. The primary aspects of this case were:

- Each agent's pool of $n_s$ pairs was combined into a single pool of $\sum_{s=\{1,\dots,4\}} n_s$ incompatible pairs.

- Both intra-agent and cross-agent matches were allowed, increasing the potential for successful transplants.

- The minimum HLA compatibility requirement was applied to all transplants to ensure quality.

- A fairness constraint was added to guarantee that each agent received at least as many transplants as they would have received independently (as in Case 1).

After reviewing the results of each case, we conducted a sensitivity analysis on the two main parameters of the problem: the number of pairs $n_s$ and the HLA compatibility threshold $L_{HLA}$. This analysis provides further insight into how these parameters affect both the quantity and quality of transplants in multi-agent kidney exchanges.

**Input Parameters**

For the base scenario, we consider the following key input parameters for the kidney exchange optimization problem:

- **Number of Pairs per Agent:** We assume that each agent has 5 patient-donor pairs.

- **Number of Agents:** The model considers 4 agents (hospitals or regions), each managing their own patient-donor pairs.

- $L_{HLA}$: This threshold value determines the minimum required HLA score for high-quality transplants. In the base scenario, we set $L_{HLA} = 210$.

- **Compatibility Constraints ($c_{ij}$):** Compatibility between a patient and donor is governed by two biological factors:

    - **Blood Type Compatibility:** The donor and recipient must have compatible blood types. Type O can donate to anyone, while Type AB can only donate to another AB.

    - **PRA Compatibility:** PRA measures the likelihood of the recipient rejecting the donor kidney. Lower PRA indicates higher compatibility.

For each patient-donor pair, we randomly generated **HLA scores** based on a predefined set of possible values: [55, 110, 150, 160, 205, 210, 255, 300, 305, 310, 350, 355, 360]. These values reflect different levels of compatibility, with higher values indicating greater compatibility between the donor and recipient (Kutlu-Gündoğdu et al., 2018). For the base case, where the **Number of Pairs per Agent = 5** and $L_{HLA} = 210$, the outcomes across the three models are summarized below. The results for Model 1, Model 2, and Model 3 are shown in Tables 2, 3, and 4, respectively.

**Table 2**
Model 1 Result in Base Scenario

| | Model 1 | | | |
|---|---|---|---|---|
| Source | Agent 1 | Agent 2 | Agent 3 | Agent 4 |
| Number of assigned kidneys | 2 | 0 | 4 | 2 |

**Table 3**
Model 2 Result in Base Scenario

| | Model 2 | | | |
|---|---|---|---|---|
| Source | Agent 1 | Agent 2 | Agent 3 | Agent 4 |
| Number of assigned kidneys | 0 | 0 | 2 | 0 |

**Table 4**
Model 3 Result in Base Scenario

| | Model 3 | | | |
|---|---|---|---|---|
| Source | Agent 1 | Agent 2 | Agent 3 | Agent 4 |
| Number of assigned kidneys | 5 | 4 | 5 | 4 |

The results are visually presented in Fig. 1, which compares the outcomes across the three models and the number of kidneys assigned to each agent.

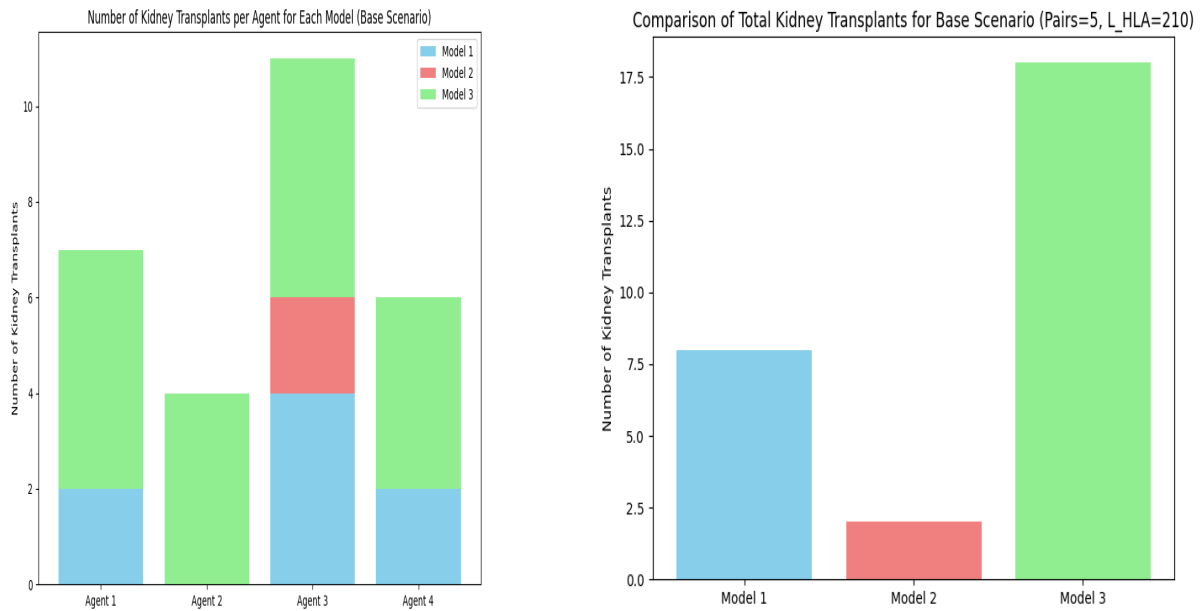

**Fig. 1.** Left plot: Comparison of models' results, Right plot: Assigned kidneys for each agent

**Interpretation of Results**

- Model 3 results in a total of 18 kidney transplants, with each agent receiving between 4 to 5 transplants. The fairness constraint ensures that no agent is disadvantaged, and the overall transplant count is significantly higher than that in Model 1 and Model 2.

- Model 1, which maximizes the number of transplants without considering HLA compatibility, yields 8 transplants, but this model only considers basic biological compatibility (blood type and PRA).

- Model 2, which enforces strict HLA compatibility through the L_HLA threshold, results in only 2 successful transplants, demonstrating the reduction in transplant numbers when prioritizing transplant quality.

These results highlight a crucial trade-off in kidney transplantation programs: when prioritizing high-quality transplants that meet stringent biological compatibility requirements, such as the $L_{HLA}$ threshold, the number of successful transplants is reduced. In Model 2, enforcing this constraint drastically reduces the number of transplants compared to Model 1, which only accounts for blood type and PRA compatibility.

However, MKEP represented by Model 3 provides a balanced solution. By pooling the incompatible pairs from all agents into a larger network, MKEPs significantly increase both the quantity and quality of transplants. Model 3 achieves the best results by not only increasing the number of successful transplants compared to individual agents working in isolation but also maintaining high-quality matches that meet the minimum HLA compatibility requirements.

This finding demonstrates the critical benefit of considering larger, more integrated kidney exchange pools. By combining multiple agents, MKEPs offer the best of both worlds—achieving both high-quality transplants and increasing the overall transplant numbers.

*4.2 Analysis*

In kidney exchange programs, the success of matching patients with compatible donors heavily depends on two key factors: the biological compatibility between patient-donor pairs and the size of the available patient-donor pool. Understanding the influence of these factors is crucial for optimizing both the number and quality of successful transplants.

*4.2.1 Sensitivity Analysis on $L_{HLA}$*

The HLA score plays a vital role in determining the success of a transplant. A higher HLA compatibility score between a patient and donor significantly reduces the likelihood of organ rejection, leading to better post-transplant outcomes. To ensure transplant quality, many kidney exchange programs enforce a minimum HLA threshold ($L_{HLA}$), which sets the acceptable HLA score for matching patient-donor pairs.

While a higher $L_{HLA}$ threshold improves transplant quality, it also reduces the number of compatible matches. This creates a trade-off between maximizing the quantity of transplants and ensuring transplant quality. Conducting a sensitivity analysis on $L_{HLA}$ helps us evaluate how varying the HLA threshold affects the overall number of successful transplants. By exploring different $L_{HLA}$ values, we can determine the optimal threshold that balances transplant quality and quantity, particularly when dealing with diverse or limited pools of patient-donor pairs. The results are presented in Table 5.

**Table 5**
Total Number of Assigned Kidneys in Each Model Based on Different Values of $L_{HLA}$

| | . Sensitivity Analysis for $L_{HLA}$ | | |
|---|---|---|---|
| $L_{HLA}$ | Model 1 (Total) | Model 2 (Total) | Model 3 (Total) |
| 205 | 8 | 4 | 20 |
| 210 | 8 | 2 | 18 |
| 215 | 8 | 2 | 18 |
| 220 | 8 | 2 | 16 |
| 225 | 8 | 0 | 16 |
| 230 | 8 | 0 | 14 |

The sensitivity analysis of the $L_{HLA}$ **threshold** reveals a clear trade-off between the quality and quantity of kidney transplants in the different models. In **Model 1**, which does not enforce any $L_{HLA}$ constraint, the number of transplants remains constant at 8, irrespective of the $L_{HLA}$ value. This model prioritizes maximizing the number of transplants without considering compatibility standards, which results in a higher number of transplants at the expense of transplant quality. On the other hand, **Model 2**, which strictly enforces the $L_{HLA}$ threshold, shows a significant reduction in the number of transplants as the threshold increases. For instance, at $L_{HLA}$ = 205, Model 2 manages 4 transplants, but this drops to 2 as $L_{HLA}$ reaches **210**, and eventually no transplants are possible at $L_{HLA}$ = **225** and beyond due to the lack of compatible pairs.

**Model 3**, which combines patient-donor pools across multiple agents, consistently outperforms both Model 1 and Model 2 by maintaining the highest number of transplants at every $L_{HLA}$ value. At $L_{HLA}$ = **205**, Model 3 achieves 20 transplants, but this number gradually decreases to 16 at $L_{HLA}$ = **220** and **225**, and further to 14 at $L_{HLA}$= **230**. The superior performance of Model 3 stems from pooling resources across agents, which increases the likelihood of finding compatible pairs even with a $L_{HLA}$ threshold. Although the number of transplants decreases as the $L_{HLA}$ threshold increases, Model 3 strikes the best balance by providing a larger number of transplants with higher compatibility, offering both high quality and quantity. This demonstrates that using a multi-agent kidney exchange system can mitigate the impact of stricter $L_{HLA}$ constraints and allow kidney exchange programs to achieve optimal results.

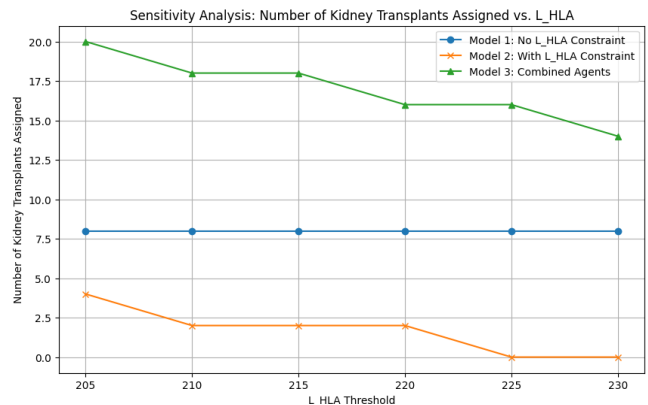

**Fig. 2.** Sensitivity Analysis on $L_{HLA}$

As shown in Fig. 2, the sensitivity analysis of $L_{HLA}$ highlights the trade-offs between transplant quality and quantity:

- **Model 1**, without the $L_{HLA}$ constraint, maximizes the number of transplants but sacrifices transplant quality.
- **Model 2**, which enforces the $L_{HLA}$ threshold, results in a significant reduction in transplants as the threshold becomes stricter.
- **Model 3** demonstrates the best balance by achieving both higher quality and quantity, especially when multiple agents are combined, as this larger pool of patients and donors provides more flexibility in matching.

*4.2.2 Sensitivity Analysis on Number of Pairs*

The size of the patient-donor pool is another critical factor that significantly influences the success of kidney exchange programs. Larger pools increase the likelihood of finding compatible matches for patients, as the number of potential combinations grows. Conversely, smaller pools limit the opportunities for successful exchanges, especially when strict biological compatibility constraints, such as **HLA**, are in place.

Conducting a sensitivity analysis on the number of pairs per agent allows us to assess how increasing pool size impacts the total number of transplants. This analysis also provides insights into whether pooling resources across multiple agents, as in **MKEPs**, can offset the limitations of smaller, individual pools. As the number of pairs per agent increases, the overall number of transplants is expected to rise, highlighting the potential benefits of expanding kidney exchange programs and optimizing resource allocation. The results are presented in Table 6.

**Table 6**
Total Number of Assigned Kidneys in Each Model Based on Different Number of Pairs

| Sensitivity Analysis for Number of Pairs | | | |
|---|---|---|---|
| Number of Pairs | Model 1 (Total) | Model 2 (Total) | Model 3 (Total) |
| 5 | 8 | 2 | 18 |
| 6 | 10 | 6 | 20 |
| 8 | 14 | 10 | 24 |
| 10 | 24 | 16 | 36 |
| 12 | 30 | 20 | 42 |

The **sensitivity analysis** on the number of pairs per agent highlights the critical role pool size plays in the success of kidney exchange programs across the three models. In **Model 1**, which does not impose an $L_{HLA}$ constraint, the number of transplants increases steadily as the pool size grows. With 5 pairs, **Model 1** achieves 8 transplants, rising to 24 with 10 pairs and 30 with 12 pairs. This demonstrates how larger pools naturally provide more matching opportunities, maximizing the number of transplants without considering biological compatibility beyond blood type and PRA.

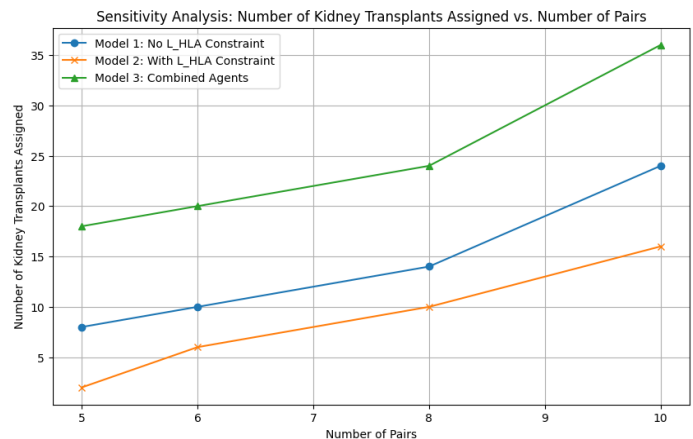

**Fig. 3.** Sensitivity Analysis on Number of Pairs

In contrast, **Model 2**, which applies the $L_{HLA}$ threshold, shows a slower increase in the number of transplants due to the stricter compatibility requirements. Starting with only 2 transplants at 5 pairs, the model reaches 16 transplants at 10 pairs and 20 transplants at 12 pairs. Despite the constraint, increasing the pool size allows **Model 2** to find more compatible pairs, highlighting the benefits of a larger pool in overcoming biological limitations such as HLA compatibility.

Model 3, which pools resources across multiple agents, consistently outperforms both Model 1 and Model 2 at every pool size, achieving the highest number of transplants. With 5 pairs, Model 3 enables 18 transplants, rising to 36 at 10 pairs and 42 at 12 pairs. The ability to pool patient-donor pairs across agents leads to significantly more matches, even under strict $L_{HLA}$

14constraints. This analysis clearly demonstrates the advantage of MKEPs, where a larger, combined pool of patients and donors allows for both higher-quality transplants and an increased number of successful matches. Expanding the size of the patient-donor pool, particularly through multi-agent coordination, is thus critical to optimizing the performance of KEP.

Using **Fig. 3**, the sensitivity analysis on the Number of Pairs per Agent clearly demonstrates the benefits of increasing the size of the patient-donor pool:

- **Model 1**, which does not apply any compatibility constraints, shows a steady increase in the number of transplants as the pool size grows.

- **Model 2**, which enforces the $L_{HLA}$ threshold, also sees improvements in the number of transplants as more pairs become available, although at a slower rate compared to Model 1, due to the stricter compatibility requirements.

- **Model 3** consistently outperforms both other models by achieving the highest number of transplants through pooling resources across agents. This highlights how **MKEPs** can overcome the limitations of smaller, individual pools, facilitating both higher transplant quality and quantity.

## 5. Conclusion

In this paper, we developed a series of mathematical models to enhance the effectiveness of KEPs by focusing on both the quantity and quality of transplants. By introducing HLA compatibility thresholds and implementing a MKEP framework, we addressed critical challenges in maximizing transplant success rates while maintaining high biological compatibility. Model 1 prioritized maximizing the number of transplants without considering HLA compatibility, leading to higher transplant numbers but compromising quality. Model 2 incorporated a minimum HLA compatibility threshold, improving transplant quality but significantly reducing the number of transplants due to stricter matching criteria. Model 3, which pooled donor-recipient pairs across multiple agents, offered the optimal solution by maximizing both the number and quality of transplants. The pooling of incompatible pairs across agents allowed for higher transplant success rates, even under the constraints of HLA compatibility, and ensured that each agent received at least as many transplants as they would have independently. Through sensitivity analyses, we demonstrated that larger patient-donor pools lead to a higher number of transplants in all models, with Model 3 consistently outperforming the others in terms of both transplant quantity and quality. The results highlight the critical trade-off between transplant quantity and quality, where stricter compatibility thresholds reduce the number of matches but improve outcomes. By leveraging multi-agent collaboration, Model 3 successfully mitigated this trade-off, providing a balanced solution that improves both the success rate and the quality of transplants. This research underscores the importance of multi-agent coordination and biological compatibility in KEP and provides valuable insights for the future of kidney transplantation. The updated version of MKEP offers a promising strategy to address the global shortage of kidney transplants, combining high standards of HLA compatibility with an expanded pool of patients and donors to increase both the number of transplants and the likelihood of long-term success. As kidney transplantation remains a critical need worldwide, these findings suggest that adopting multi-agent systems and focusing on compatibility can significantly improve the efficiency and outcomes of kidney exchange programs, ultimately benefiting both patients and healthcare systems.

## 6. Future research

While this study presents significant advancements in optimizing kidney exchange programs through multi-agent collaboration and incorporating HLA compatibility, several avenues for future research remain. One promising direction is the exploration of more sophisticated algorithms that can handle even larger pools of incompatible donor-recipient pairs, especially in real-time kidney exchange platforms. Additionally, integrating other biological compatibility measures, such as genetic matching beyond HLA, could further improve transplant success rates. Another area of potential research involves expanding the multi-agent framework to international kidney exchange programs, allowing for cross-border transplants, which could drastically increase the pool size and matching opportunities.